\documentclass[conference]{IEEEtran}
\IEEEoverridecommandlockouts

\usepackage{cite}

\usepackage{amsmath,amssymb,amsfonts}
\usepackage{algorithmic}
\usepackage{graphicx}
\usepackage{textcomp}
\usepackage{xcolor}
\usepackage{booktabs} 
\usepackage{tikz}
\usepackage{pgfplots}
\usepgfplotslibrary{fillbetween}
\def\BibTeX{{\rm B\kern-.05em{\sc i\kern-.025em b}\kern-.08em
    T\kern-.1667em\lower.7ex\hbox{E}\kern-.125emX}}
\begin{document}

\title{Rectifying Geometric Misalignment: Online Source-Free Adaptation for Class-Imbalanced EEG}

\author{\IEEEauthorblockN{Shiwen Chu$^{1,2}$, Shanglin Li$^{1,3}$, Motoaki Kawanabe$^{1,2}$, Reinmar Kobler$^{1}$}
\IEEEauthorblockA{\textit{$^{1}$Advanced Telecommunications Research Institute International, 2-2-2 Hikaridai Seika-cho, Soraku-gun, Kyoto, Japan} \\
\textit{$^{2}$Nara Institute of Science and Technology, 8916-5 Takayama-cho, Ikoma, Nara, Japan}\\
\textit{$^{3}$BIFOLD – Berlin Institute for the Foundations of Learning and Data, 10587 Berlin, Germany}\\
kawanabe@atr.jp, kobler.reinmar@gmail.com}
}

\maketitle

\begin{abstract}
Electroencephalography (EEG) based Brain-Computer Interfaces (BCIs) often require unsupervised domain adaptation (UDA) to generalize across subjects and sessions. While Riemannian alignment methods like the Riemannian Centering Transformation (RCT) are effective for handling covariate shifts, they implicitly assume balanced class priors. However, in realistic online BCI scenarios, the label distributions vary dynamically (label shift), causing standard alignment techniques to geometrically misalign the target data distributions. In this work, we propose \textbf{OSPDIM} (Online SPD manifold information maximization), a source-free online UDA framework designed to address label shifts on the Riemannian manifold. OSPDIM introduces a manifold-constrained bias parameter into the tangent space mapping, which is optimized via information maximization to correct the geometric skew caused by imbalanced data streams. Unlike offline methods relying on global batch statistics, OSPDIM estimates and corrects geometric bias on-the-fly. Simulations on $2 \times 2$ SPD matrices visually demonstrate that OSPDIM successfully rectifies the misalignment where standard centering fails. Extensive experiments on multiple motor imagery datasets show that OSPDIM significantly outperforms standard Riemannian baselines, particularly in challenging online adaptation scenarios with severe class imbalance, offering a robust solution for practical, plug-and-play BCI systems.
\end{abstract}

\begin{IEEEkeywords}
BCI, Riemannian Geometry, Online Learning, Source-Free Adaptation, Label Shift
\end{IEEEkeywords}

\section{Introduction}

EEG-based Brain-Computer Interfaces (BCIs) are promising for rehabilitation but suffer from severe domain shifts due to non-stationarity and subject variability \cite{WOLPAW2002767}. To mitigate these shifts without resource-intensive calibration \cite{Lotte2018}, Unsupervised Domain Adaptation (UDA) has emerged as a standard solution to transfer knowledge from labeled source domains to unlabeled targets \cite{Wu2020,7379089}.

Developing BCIs that generalize across domains without labeled calibration data remains a grand challenge \cite{Fairclough2020, Wolpaw2002}, corresponding to source-free unsupervised domain adaptation (SFUDA) when target data are unavailable during training \cite{Liu2021, Yang2021}. Riemannian geometry-aware alignment methods for SPD matrix-valued EEG features are strong candidates for this setting \cite{ju2026spdmatrixlearningneuroimaging, Barachant2011, Roy2022, Mellot2023}, as they provide interpretable and robust covariance-based representations \cite{Congedo2017, Sabbagh2020, Kobler2021}. These methods mainly align Riemannian moments, such as the Fr\'echet mean and variance, so that a source-trained model can generalize to related target domains \cite{Zanini2017, Rodrigues2019, li2026heegnet,Kobler2022a, Wei2022, Ju2022, Mellot2024, li2024geometric}.

Despite the progress in alignment techniques, a critical but frequently overlooked issue is label shift, which appears in many real-world applications such as EEG-based sleep staging \cite{Tholke2023}. In online BCI use, such shifts can arise because user intentions and mental states are not sampled uniformly over time: some commands may be issued more frequently, and fatigue, attention fluctuations, or task interruptions may produce stream segments dominated by only a subset of classes. Under conditions of label shift, aligning the moments of marginal feature distributions can be detrimental \cite{Bakas2023}. Consequently, addressing the diverse sources of distribution shifts in EEG requires an SFUDA approach capable of explicitly handling label shifts. Although the broader machine learning literature offers several frameworks for SFUDA under label shift conditions \cite{Li2021_MM, Liang2024}, few have been rigorously applied to EEG data. For example, Li et al.~\cite{Li2023} employed the objective of information maximization \cite{Shi2012} for cross-domain generalization. Within the realm of Riemannian geometry methods, Mellot et al.~\cite{Mellot2024} studied an EEG-based age regression problem and proposed a framework to facilitate generalization between populations with different prior distributions.

Here, we address the critical challenge of online label shift in EEG-based BCIs, where standard Riemannian alignment fails catastrophically due to the geometric misalignment of streaming statistics. We extend the SPDIM framework \cite{SPDIM} to the challenging online source-free setting. Unlike the original offline formulation that relies on global batch statistics, we propose a novel sequential adaptation protocol designed to estimate and correct the geometric bias on-the-fly. This mechanism employs a sliding buffer and iterative manifold optimization to dynamically rectify the alignment as data streams in. Furthermore, we provide a comprehensive analysis of adaptation speed and stability, demonstrating that Online SPDIM significantly outperforms standard baselines even under severe label shift.

\section{Preliminaries}
\subsection{Riemannian geometry}
The smooth manifold of real $D \times D$ symmetric positive definite (SPD) matrices, denoted as $\mathcal{S}_D^+ = \{Z \in \mathbb{R}^{D \times D} : Z^T = Z, Z \succ 0\}$, forms a Riemannian manifold when equipped with an inner product on the tangent space $\mathcal{T}_Z \mathcal{S}_D^+$ at each point $Z$. In this work, we employ the Affine Invariant Riemannian Metric (AIRM) as the inner product. The tangent spaces possess a Euclidean structure with computationally efficient distances, locally approximating the Riemannian distances on $\mathcal{S}_D^+$. 
The Riemannian distance between two SPD matrices $A, B \in \mathcal{S}_D^+$ is defined as $\delta_R(A, B) = \|\log(A^{-1/2}BA^{-1/2})\|_F$. For a set of SPD points $\mathcal{Z} = \{Z_i\}_{i=1}^n \subset \mathcal{S}_D^+$, the Fr\'echet mean is defined as the unique point that minimizes the sum of squared Riemannian distances:
\begin{equation}
    G_{\mathcal{Z}} = \arg\min_{M \in \mathcal{S}_D^+} \sum_{i=1}^n \delta_R^2(M, Z_i).
\end{equation}

\subsection{Class-imbalanced online unsupervised domain adaptation}

\textbf{Problem Setup.} 
Let $x$ denote the input data and $y$ the corresponding output labels. We consider a set of $N$ source domains $\{\mathcal{D}_{s_i}\}_{i=1}^N$ and $M$ target domains $\{\mathcal{D}_{t_j}\}_{j=1}^M$, with marginal probability distributions $p_{s_i}(x)$ and $q_{t_j}(x)$, respectively. Each source domain $\mathcal{D}_{s_i} = \{(x_{s_i}^{(k)}, y_{s_i}^{(k)})\}_{k=1}^{l_{s_i}}$ consists of $l_{s_i}$ labeled examples, while each target domain $\mathcal{D}_{t_j} = \{x_{t_j}^{(k)}\}_{k=1}^{l_{t_j}}$ contains $l_{t_j}$ unlabeled examples.

We operate under specific distributional assumptions to model the domain shift. First, we assume \textit{covariate shift} with the input marginal distributions differing between domains (i.e., $p_{s_i}(x) \neq q_{t_j}(x)$). Second, we model the \textit{class-imbalance} problem by assuming \textit{label shift} between the source and target domains (i.e., $p_{s_i}(y) \neq q_{t_j}(y)$), while the source label distributions are consistent (i.e., $p_{s_i}(y) = p_{s_k}(y)$). Furthermore, consistent with the Riemannian alignment framework, we assume that the distribution shifts manifest primarily as geometric displacements of the Riemannian Fr\'echet mean, which can be corrected via geometric re-centering.

The primary objective is to transfer knowledge from the source domains $\{\mathcal{D}_{s_i}\}_{i=1}^N$ to the target domains $\{\mathcal{D}_{t_j}\}_{j=1}^M$ and learn a robust target prediction function $h_t : x_t \to y_t$, utilizing only the labeled source data and the unlabeled target data.

\textbf{Online setting.} 
In contrast to traditional offline adaptation where the entire target dataset $\mathcal{D}_{t_j}$ is accessible simultaneously, we consider a strictly \textit{online} unsupervised domain adaptation scenario. Here, the target data is revealed sequentially in a stream or as mini-batches. At any time step $k$, the model has only access to the current observation (or batch) $x_{t_j}^{(k)}$ to update its parameters and predict the label. Crucially, the model cannot access future data and does not retain a large buffer of historical data due to memory limitations and computational efficiency requirements. This setting necessitates an alignment and prediction mechanism capable of adapting efficiently on-the-fly to the shifting target distribution.

\section{Methods}

\subsection{TSMNet and Geometric Misalignment}

Standard Tangent Space Mapping (TSM) frameworks \cite{Barachant2011} project SPD covariances into a Euclidean tangent space centered at the Riemannian Fr\'echet mean. TSMNet \cite{Kobler2022a} extends this by integrating domain-specific batch normalization into an end-to-end architecture, which aligns marginal distributions via geometric re-centering and whitening. While this mechanism effectively handles covariate shift, it is highly susceptible to label shift. Specifically, aligning the biased empirical mean of an imbalanced target stream to the identity matrix (the balanced source center) introduces severe geometric distortion, thereby misaligning class clusters and degrading performance.

\subsection{Online SPD Manifold Information Maximization (OSPDIM)}
\label{sec:ospdim}

Standard Riemannian alignment methods, such as the Riemannian Centering Transformation (RCT)\cite{Zanini2017} incorporated in TSMNet, operate on the assumption that label distributions remain consistent across domains. However, in the presence of label shift, aligning the Fr\'echet means of marginal distributions leads to geometric over-correction. To address this in real-time scenarios, we propose \textbf{OSPDIM}, which extends the offline SPDIM framework \cite{SPDIM} to the challenging online setting.

\textbf{Manifold-constrained Bias Parameter.} 
OSPDIM operates in a source-free setting, keeping the pre-trained source feature extractor $f_\theta$ and classifier $g_\psi$ frozen. To counteract the geometric misalignment during test-time adaptation, we introduce a time-varying domain-specific bias parameter $\Phi_t \in \mathcal{S}_D^+$ into the tangent space mapping. The modified mapping function for an incoming target sample $C_i$ at time $t$ is defined as:
\begin{equation}
    m_{\phi}(C_i) = \text{upper} \circ \log\left(\Phi_t^{\frac{1}{2}} \left( \overline{C}_t^{-\frac{1}{2}} C_i \overline{C}_t^{-\frac{1}{2}} \right) \Phi_t^{\frac{1}{2}}\right)
\end{equation}
where $\overline{C}_t$ represents the reference center for tangent space mapping at time $t$. Intuitively, the inner term performs standard Riemannian centering (whitening) relative to the empirical target mean. The outer multiplication applies a learnable geometric translation on the SPD manifold. In standard TSMNet, $\overline{C}_t$ is a fixed batch statistic and no bias is applied ($\Phi_t=I$). In OSPDIM, we initialize $\Phi_t$ as the identity but dynamically optimize it to minimize the information maximization loss (see Eq. \ref{eq:im}), thereby actively recovering the true class-balanced geometric center.

\textbf{Online Adaptation Protocol.} 
The core contribution of OSPDIM is the sequential adaptation mechanism. Unlike the offline setting where bias is optimized globally, OSPDIM employs a sliding window approach:
\begin{itemize}
    \item \textbf{Sliding Buffer:} We maintain a First-In-First-Out (FIFO) buffer $\mathcal{B}_t$ of capacity $N$ to store the most recent target features. This buffer stabilizes the gradient estimation against single-sample noise.
    \item \textbf{Iterative Manifold Update:} At each time step $t$, we update the bias $\Phi_t$ to minimize the loss calculated on the target buffer $\mathcal{B}_t$. Since $\Phi_t$ resides on the Riemannian manifold $\mathcal{S}_D^+$, standard Euclidean updates are invalid. We employ approximated Riemannian gradient descent via Euclidean optimization followed by a retraction:
    \begin{equation}
        \Phi_{t} \leftarrow \text{Proj}_{\mathcal{S}_D^+} \left( \Phi_{t-1} - \eta \cdot \nabla_{\Phi} \mathcal{L}_{IM}(\mathcal{B}_t; \Phi_{t-1}) \right)
    \end{equation}
    where $\nabla_{\Phi}$ is the Euclidean gradient computed by Adam, and $\text{Proj}_{\mathcal{S}_D^+}$ acts as a retraction mapping the updated parameter back onto the SPD manifold to ensure validity. This allows OSPDIM to track and correct non-stationary label shifts in real-time.
\end{itemize}

\textbf{Optimization via Information Maximization.} 
To estimate $\Phi_t$ without target labels, we optimize the information maximization (IM) objective on the buffer data. Let $\hat{p}_i = g_\psi(m_{\phi_t}(C_i))$ denote the $K$-dimensional softmax probability predicted by the frozen source model for an unlabeled target sample $C_i \in \mathcal{B}_t$. The loss function $\mathcal{L}_{IM}$ is defined as:
\begin{equation}
\label{eq:im}
    \mathcal{L}_{IM} = \underbrace{-\frac{1}{|\mathcal{B}_t|} \sum_{C_i \in \mathcal{B}_t} \sum_{c=1}^K \hat{p}_{i,c} \log \hat{p}_{i,c}}_{\mathcal{L}_{CEM}} + \underbrace{\sum_{c=1}^K \bar{p}_c \log \bar{p}_c}_{\mathcal{L}_{MEM}}
\end{equation}
where $\bar{p}_c = \frac{1}{|\mathcal{B}_t|} \sum_{C_i \in \mathcal{B}_t} \hat{p}_{i,c}$ is the mean prediction probability for class $c$ across the buffer. The Conditional Entropy Minimization ($\mathcal{L}_{CEM}$) acts on individual target predictions to encourage confidence, pushing decision boundaries away from high-density regions. The Marginal Entropy Maximization ($\mathcal{L}_{MEM}$) acts on the buffer-level label distribution to prevent trivial solutions (e.g., assigning all samples to a single class) by encouraging diverse predictions.
 
\section{Experiments}

\subsection{Simulations}

\begin{figure*}[t]
    \centering
    \includegraphics[width=\textwidth]{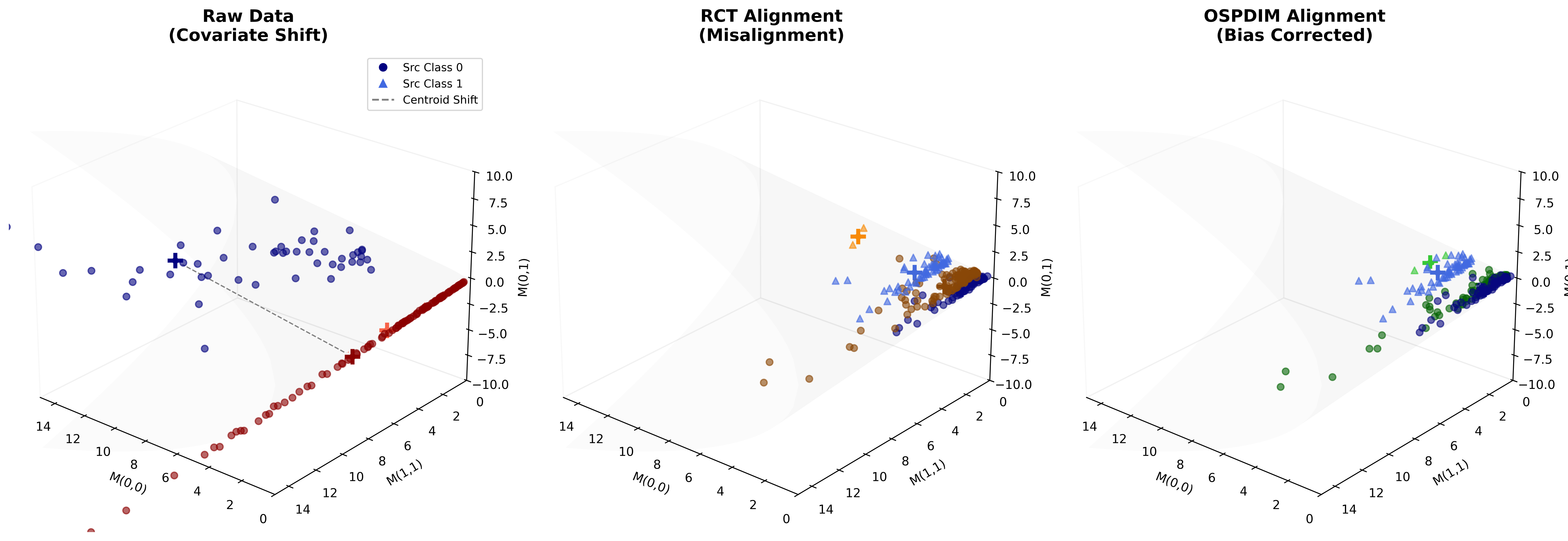} 
    \caption{\textbf{Visualization of domain adaptation on the $2 \times 2$ SPD manifold.} 
    The gray curved structure represents the SPD manifold. 
    \textbf{Left (Raw Data):} Blue dots represent Source, Red dots represent Target. Class centroids are marked by `+'. The dashed line connects the majority class centroids, highlighting the initial shift. Note: markers (circles/triangles) distinguish class labels.
    \textbf{Middle (RCT):} Standard alignment fails under label shift. Target data (Orange) is centered to the identity matrix $I$, but the class clusters remain misaligned with the source, as shown by the distinct black gap between class centroids.
    \textbf{Right (OSPDIM):} Our method (Green dots) effectively reduces the gap between source and target centroids by optimizing the manifold-constrained bias, accurately recovering the decision boundary structure.}
    \label{fig:sim_viz}
\end{figure*}

To validate our approach, we visualized the adaptation process on the $2 \times 2$ SPD manifold. We generated latent features and mapped them to SPD matrices using domain-specific mixing matrices, resulting in ill-conditioned, "needle-like" distributions characteristic of EEG signals (Figure \ref{fig:sim_viz}, Left). To simulate severe label shift, the target domain was subsampled to a minority-to-majority ratio of 0.1.

As shown in Figure \ref{fig:sim_viz} (Middle), standard RCT effectively centers the global distribution around the Identity matrix. However, due to the severe class imbalance, the global mean is dominated by the majority class. Consequently, simply aligning the global means results in \textbf{semantic misalignment}: the centroid of the target majority class (Orange `+') is pulled away from the source majority class (Blue `+'), as indicated by the distinct black connecting line.

In contrast, \textbf{OSPDIM} (Right) optimizes a manifold-constrained bias parameter via information maximization. This effectively corrects the geometric skew caused by the label shift. As observed, the centroids of the corresponding classes between the source and target domains (e.g., Source Class 0 and Target Class 0) become nearly overlapping, effectively realigning the target distribution (Green) with the source. This visual confirmation demonstrates that OSPDIM successfully recovers the true class distribution structure without requiring target labels.

\subsection{Motor imagery}

To validate the proposed framework in realistic BCI scenarios, we evaluated our method on public motor imagery datasets. While standard benchmarks are typically balanced, real-world online BCI applications often encounter varying class priors over time. To bridge this gap, we artificially introduced label shifts in the target domains to simulate challenging online adaptation scenarios.

\textbf{Datasets and Preprocessing.}
We utilized two widely used motor imagery datasets from the MOABB framework and BCI competition IV \cite{BCICompetitionIV}: BNCI2014001 (9 subjects, 2 classes, 144 trials per class) and BNCI2015001 (12 subjects, 2 classes, 200 trials per class). The preprocessing pipeline included resampling EEG signals to 256 Hz and applying temporal band-pass filters (4--36 Hz) to capture sensorimotor rhythms. To ensure data quality, we performed outlier rejection by removing trials with peak-to-peak amplitude exceeding 200 $\mu$V.  We extracted task-related epochs associated with specific class labels for classification, using a time window of [0.5, 3.5] seconds post-cue to capture the event-related desynchronization effects. Regarding the problem formulation, we assume that while the source domains (different subjects) exhibit covariate shift, their label distributions $P_S(y)$ remain balanced and consistent, as is standard in controlled BCI calibration protocols.

\textbf{Experimental Setup.}
We adopted a strict source-free unsupervised domain adaptation (SFUDA) protocol. For each dataset, we employed a leave-one-subject-out cross-validation scheme. The source model (TSMNet) was trained on labeled source subjects and then frozen. Source data is used only during pre-training; during online adaptation, only the frozen source model and incoming unlabeled target samples are available. For the target subject, we simulated varying degrees of label shift by subsampling the data to achieve minority-to-majority class ratios (Imbalance Ratio) of $\{0.1, 0.2, 0.3, 0.4\}$. We focus on this range because the goal is to stress-test online adaptation under severe-to-moderate imbalance, where biased Riemannian centering is expected to be most harmful. Higher ratios such as $\rho=0.9$ are close to the balanced case and therefore induce only a weak geometric bias, making them less central to the failure mode studied here. To increase statistical power and mitigate the noise introduced by the random subsampling process, we run each experiment across 10 different random seeds for every subject and imbalance ratio, and we report the average gain per subject.

We compared five settings: \textbf{OSPDIM (Proposed)}, our online method using a sliding buffer ($N=32$) and $K=5$ update steps per sample; \textbf{SPDIM (Offline Reference)} \cite{SPDIM}, an offline upper bound using the full target batch; \textbf{Deep Learning}, an EEGConformer trained on source domains without adaptation, serving as a non-geometric reference baseline; \textbf{Online RCT}, which updates the target mean sequentially via a Riemannian exponential moving average, serving as a standard geometric baseline for streaming data; and \textbf{Offline RCT}, which computes the target Fr\'echet mean from the entire target set.

\begin{figure*}[t]
    \centering
    \includegraphics[width=0.8\textwidth]{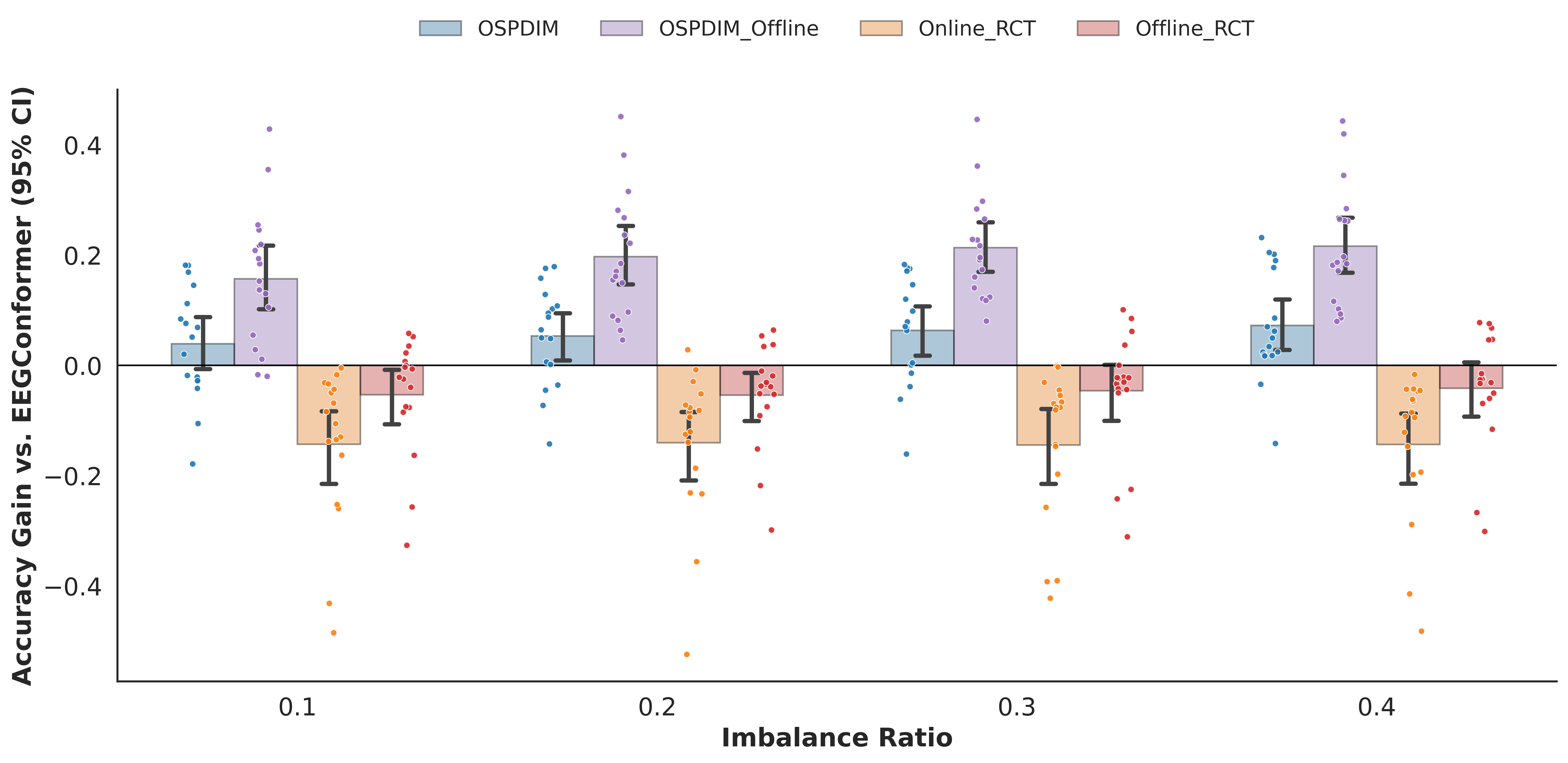}
    \caption{\textbf{Performance Gain Analysis.} 
    Comparison of balanced accuracy gains relative to the deep learning baseline (EEGConformer, zero line). 
    Each dot represents the average gain of a single subject across multiple random seeds. 
    Bars indicate the group-level mean gain with 95\% confidence intervals.
    OSPDIM consistently achieves positive gains across all imbalance ratios on both datasets, demonstrating robust correction of geometric misalignment. }
    \label{fig:gain_analysis}
\end{figure*}

\textbf{Results.}
The quantitative results are summarized in Figure \ref{fig:gain_analysis}, which reports the balanced accuracy gain relative to the deep learning baseline. We observed distinct performance patterns across the datasets:

\textit{Sensitivity of RCT to Label Shift}
Consistent with the simulations, RCT baselines show substantial performance losses under label shift. The degradation remains relatively stable across $\rho \in [0.1,0.4]$, suggesting that even mild imbalance can move the target mean enough to distort the source decision boundary; stronger imbalance further increases centroid bias but does not necessarily cause a proportional accuracy drop.

\textit{Effectiveness of OSPDIM.}
OSPDIM consistently achieves positive gains across both datasets and all imbalance ratios. Subject-level averages and bootstrap-based 95\% confidence intervals ($n=2000$) indicate consistent improvements across subjects, and it retained stable performance when label shift is absent, showing that the learned manifold-constrained bias recovers performance lost to domain shift.

\textit{Performance Gap and Limitations}
OSPDIM remains below the offline SPDIM upper bound because it relies only on a small causal buffer rather than full target-batch statistics. Nevertheless, it bridges much of the gap between failing Online RCT and the offline reference, supporting the effectiveness of information-maximization-based bias correction in imbalanced streams.

\subsection{Ablation Study: Adaptation Speed}
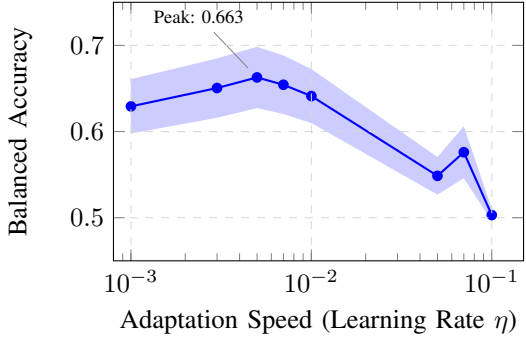
\begin{figure}[t]
\centering
\begin{tikzpicture}
\begin{semilogxaxis}[
    width=7cm, height=5cm,
    xlabel={Adaptation Speed (Learning Rate $\eta$)},
    ylabel={Balanced Accuracy},
    grid=major,
    grid style={dashed,gray!30},
    xmin=0.0008, xmax=0.15,
    ymin=0.45, ymax=0.75,
    legend pos=north east,
    legend style={nodes={scale=0.8, transform shape}},
    axis on top, 
]

\addplot[name path=mean, draw=none] coordinates {
    (0.001, 0.6292) (0.003, 0.6505) (0.005, 0.6628) (0.007, 0.6543) 
    (0.010, 0.6412) (0.050, 0.5486) (0.070, 0.5760) (0.100, 0.5031)
};

\addplot[name path=upper, draw=none] coordinates {
    (0.001, 0.6608) (0.003, 0.6849) (0.005, 0.6984) (0.007, 0.6883) 
    (0.010, 0.6725)  (0.050, 0.5704) (0.070, 0.6061) (0.100, 0.5102)
};

\addplot[name path=lower, draw=none] coordinates {
    (0.001, 0.5977) (0.003, 0.6160) (0.005, 0.6273) (0.007, 0.6204) 
    (0.010, 0.6099) (0.050, 0.5268) (0.070, 0.5459) (0.100, 0.4960)
};

\addplot[blue!20] fill between[of=upper and lower];
\addlegendimage{area legend, blue!20, fill=blue!20} 

\addplot[thick, color=blue, mark=*, mark options={fill=blue, scale=0.8}] coordinates {
    (0.001, 0.6292) (0.003, 0.6505) (0.005, 0.6628) (0.007, 0.6543) 
    (0.010, 0.6412) (0.050, 0.5486) (0.070, 0.5760) (0.100, 0.5031)
};

\node[pin=95:{\scriptsize Peak: 0.663}, color=red] at (axis cs:0.005, 0.6628) {};

\end{semilogxaxis}
\end{tikzpicture}
\caption{\textbf{Sensitivity Analysis of Adaptation Speed.} The solid blue line represents the mean balanced accuracy of OSPDIM, while the shaded area indicates the standard error (SE).}
\label{fig:ablation_shaded}
\end{figure}

We further investigated the impact of the adaptation speed, controlled by the learning rate $\eta$ in the iterative update step. Figure \ref{fig:ablation_shaded} illustrates the performance on BNCI2014001 (Ratio 0.1) under varying $\eta$.
Small learning rates ($\eta<0.003$) update the bias too slowly, whereas large rates ($\eta>0.03$) make the bias unstable under small-buffer noise. OSPDIM reaches its best performance around $\eta \approx 0.005$ ($\sim$66\%), suggesting a trade-off between stability and responsiveness.
This confirms that tuning the adaptation speed is crucial for handling dynamic label shifts in online BCIs.

\section{Conclusion}

In this work, we presented OSPDIM, an online source-free domain adaptation framework designed for EEG decoding under non-stationary label shifts. Our simulations on the SPD manifold elucidated the geometric failure of standard Riemannian alignment (RCT): under severe class imbalance, the empirical target mean becomes biased, causing RCT to inadvertently force class clusters away from the optimal decision boundaries. OSPDIM addresses this by introducing a sequential, manifold-constrained bias correction optimized via information maximization. This approach decouples geometric centering from distribution alignment, effectively recovering the true class structure on-the-fly.

Empirical results demonstrate that OSPDIM significantly outperforms standard online baselines (e.g., $>15\%$ improvement on BNCI2014001), showing strong resilience to extreme imbalance without requiring target labels. While the strict online constraint introduces a performance gap compared to the offline oracle, OSPDIM provides a robust foundation for plug-and-play BCI systems. Future work will focus on integrating this geometric bias correction directly into end-to-end deep learning architectures to further enhance representation learning from raw EEG signals.


{
\small
\bibliographystyle{IEEEtran}
\bibliography{references}

@article{Lotte2018,
  author  = {Fabien Lotte and Laurent Bougrain and Andrzej Cichocki and others},
  title   = {A review of classification algorithms for {EEG}-based brain--computer interfaces: a 10 year update},
  journal = {Journal of Neural Engineering},
  year    = {2018},
}

@article{Wu2020,
  author  = {Dongrui Wu and Yifei Xu and Bao-Liang Lu},
  title   = {Transfer learning for {EEG}-based brain--computer interfaces: A review of progress made since 2016},
  journal = {IEEE Transactions on Cognitive and Developmental Systems},
  year    = {2020},
}

@article{Fairclough2020,
  author  = {Stephen H. Fairclough and Fabien Lotte},
  title   = {Grand challenges in neurotechnology and system neuroergonomics},
  journal = {Frontiers in Neuroergonomics},
  year    = {2020},
}

@article{Wolpaw2002,
  author  = {Jonathan R. Wolpaw and Niels Birbaumer and Dennis J. McFarland and others},
  title   = {Brain--computer interfaces for communication and control},
  journal = {Clinical Neurophysiology},
  year    = {2002},
}

@inproceedings{Liu2021,
  author    = {Yuang Liu and Wei Zhang and Jun Wang},
  title     = {Source-free domain adaptation for semantic segmentation},
  booktitle = {CVPR},
  year      = {2021},
}

@inproceedings{Yang2021,
  author    = {Shiqi Yang and Yaxing Wang and Joost Van De Weijer and others},
  title     = {Generalized source-free domain adaptation},
  booktitle = {ICCV},
  year      = {2021},

}

@article{Barachant2011,
  author  = {Alexandre Barachant and Stephane Bonnet and Marco Congedo and Christian Jutten},
  title   = {Multiclass brain--computer interface classification by {R}iemannian geometry},
  journal = {IEEE Transactions on Biomedical Engineering},
  year    = {2011},
}

@article{Roy2022,
  author  = {Raphaelle N. Roy and Marcel F. Hinss and Ludovic Darmet and others},
  title   = {Retrospective on the First Passive Brain-Computer Interface Competition on Cross-Session Workload Estimation},
  journal = {Frontiers in Neuroergonomics},
  year    = {2022},
 
}

@article{Mellot2023,
  author  = {Apolline Mellot and Antoine Collas and Pedro L. C. Rodrigues and others},
  title   = {Harmonizing and aligning {M/EEG} datasets with covariance-based techniques to enhance predictive regression modeling},
  journal = {Imaging Neuroscience},
  year    = {2023},
 
}

@article{Congedo2017,
  author  = {Marco Congedo and Alexandre Barachant and Rajendra Bhatia},
  title   = {{R}iemannian geometry for {EEG}-based brain-computer interfaces; a primer and a review},
  journal = {Brain-Computer Interfaces},
  year    = {2017},
}

@article{Sabbagh2020,
  author  = {David Sabbagh and Pierre Ablin and Ga{\"e}l Varoquaux and others},
  title   = {Predictive regression modeling with {MEG/EEG}: from source power to signals and cognitive states},
  journal = {NeuroImage},
  year    = {2020},

}

@inproceedings{Kobler2021,
  author    = {Reinmar J. Kobler and Jun-Ichiro Hirayama and Lea Hehenberger and Catarina Lopes-Dias and Gernot R. M{\"u}ller-Putz and Motoaki Kawanabe},
  title     = {On the interpretation of linear {R}iemannian tangent space model parameters in {M/EEG}},
  booktitle = {EMBC},
  year      = {2021},
}

@article{Zanini2017,
  author  = {Paolo Zanini and Marco Congedo and Christian Jutten and others},
  title   = {Transfer learning: A {R}iemannian geometry framework with applications to brain--computer interfaces},
  journal = {IEEE Transactions on Biomedical Engineering},
  year    = {2017},

}

@article{Rodrigues2019,
  author  = {Pedro Luiz Coelho Rodrigues and Christian Jutten and Marco Congedo},
  title   = {{R}iemannian {P}rocrustes Analysis: Transfer Learning for Brain--Computer Interfaces},
  journal = {IEEE Transactions on Biomedical Engineering},
  year    = {2019},
 
}

@inproceedings{Kobler2022a,
  author    = {Reinmar Kobler and Jun-ichiro Hirayama and Qibin Zhao and Motoaki Kawanabe},
  title     = {{SPD} domain-specific batch normalization to crack interpretable unsupervised domain adaptation in {EEG}},
  booktitle = {NeurIPS},
  year      = {2022},
 
}

@inproceedings{Wei2022,
  author    = {Xiaoxi Wei and A. Aldo Faisal and Moritz Grosse-Wentrup and others},
  title     = {2021 {BEETL} Competition: Advancing Transfer Learning for Subject Independence and Heterogenous {EEG} Data Sets},
  booktitle = {NeurIPS},
  year      = {2022},
 
}

@article{Ju2022,
  author  = {Ce Ju and Cuntai Guan},
  title   = {Deep Optimal Transport on {SPD} Manifolds for Domain Adaptation},
  journal = {arXiv preprint arXiv:2201.05745},
  year    = {2022}
}

@article{Mellot2024,
  author  = {Apolline Mellot and Antoine Collas and Sylvain Chevallier and others},
  title   = {Geodesic optimization for predictive shift adaptation on {EEG} data},
  journal = {arXiv preprint arXiv:2407.03878},
  year    = {2024}
}

@article{Tholke2023,
  author  = {Philipp Tholke and Yorguin-Jose Mantilla-Ramos and Hamza Abdelhedi and others},
  title   = {Class imbalance should not throw you off balance: Choosing the right classifiers and performance metrics for brain decoding with imbalanced data},
  journal = {NeuroImage},
  year    = {2023},
}

@article{Bakas2023,
  author  = {Stylianos Bakas and Siegfried Ludwig and Dimitrios A. Adamos and others},
  title   = {Latent alignment with deep set {EEG} decoders},
  journal = {arXiv preprint arXiv:2311.17968},
  year    = {2023}
}

@inproceedings{Li2021_MM,
  author    = {Xinhao Li and Jingjing Li and Lei Zhu and Guoqing Wang and Zi Huang},
  title     = {Imbalanced source-free domain adaptation},
  booktitle = {ACMMM},
  year      = {2021},
}

@article{Liang2024,
  author  = {Jian Liang and Ran He and Tieniu Tan},
  title   = {A Comprehensive Survey on Test-Time Adaptation Under Distribution Shifts},
  journal = {International Journal of Computer Vision},
  year    = {2024},
}

@article{Li2023,
  author  = {Siyang Li and Ziwei Wang and Hanbin Luo and Lieyun Ding and Dongrui Wu},
  title   = {T-time: Test-time information maximization ensemble for plug-and-play {BCIs}},
  journal = {IEEE Transactions on Biomedical Engineering},
  year    = {2023},
}

@inproceedings{Shi2012,
  author    = {Yuan Shi and Fei Sha},
  title     = {Information-theoretical learning of discriminative clusters for unsupervised domain adaptation},
  booktitle = {ICML},
  year      = {2012},
  pages     = {1275--1282}
}

@article{BCICompetitionIV,
author={Tangermann, Michael  and Müller, Klaus-Robert  and Aertsen, Ad  and others},  
TITLE={Review of the {BCI} Competition {IV}},    
JOURNAL={Frontiers in Neuroscience},      
VOLUME={Volume 6 - 2012},
YEAR={2012},

}

@misc{SPDIM,
      title={{SPDIM}: Source-Free Unsupervised Conditional and Label Shift Adaptation in {EEG}}, 
      author={Shanglin Li and Motoaki Kawanabe and Reinmar J. Kobler},
      year={2024},
}

@article{WOLPAW2002767,
title = {Brain–computer interfaces for communication and control},
journal = {Clinical Neurophysiology},
year = {2002},
author = {Jonathan R Wolpaw and Niels Birbaumer and Dennis J McFarland and others},
}

@ARTICLE{7379089,
  author={Jayaram, Vinay and Alamgir, Morteza and Altun, Yasemin and others},
  journal={IEEE Computational Intelligence Magazine}, 
  title={Transfer Learning in Brain-Computer Interfaces}, 
  year={2016},
}

@misc{ju2026spdmatrixlearningneuroimaging,
      title={{SPD} Matrix Learning for Neuroimaging Analysis: Perspectives, Methods, and Challenges}, 
      author={Ce Ju and Reinmar Kobler and Antoine Collas and Motoaki Kawanabe and others},
      year={2026},

}

@article{li2026heegnet,
  title={HEEGNet: Hyperbolic Embeddings for EEG},
  author={Li, Shanglin and Chu, Shiwen and Ko{\c{c}}, Okan and Ding, Yi and Zhao, Qibin and Kawanabe, Motoaki and Chen, Ziheng},
  journal={arXiv preprint arXiv:2601.03322},
  year={2026}
}

@inproceedings{li2024geometric,
  title={Geometric Deep Learning to Enhance Imbalanced Domain Adaptation in EEG.},
  author={Li, Shanglin and Kawanabe, Motoaki and Kobler, Reinmar J},
  booktitle={ESANN},
  year={2024}
}
}

\end{document}